\documentclass[preprint,3p,authoryear]{elsarticle}

\usepackage{amssymb}
\usepackage{booktabs}
\usepackage{multirow}
\usepackage{subcaption}
\usepackage{textcomp,mathcomp}
\usepackage[table]{xcolor}
\usepackage{hyperref}
\hypersetup{colorlinks=true,allcolors=blue}

\usepackage{amsthm}
\usepackage{amsmath}

\journal{REC 2024}

\begin{document}

\begin{frontmatter}



\title{Towards robust prediction of material properties for nuclear reactor design under scarce data - a study in creep rupture property}


\author[inst1]{Yu Chen}

\affiliation[inst1]{organization={Centre for Intelligent Infrastructure, University of Strathclyde},
            city={Glasgow},
            country={UK}}

\author[inst1]{Edoardo Patelli}
\author[inst1]{Zhen Yang}
\author[inst2]{Adolphus Lye}

\affiliation[inst2]{organization={Singapore Nuclear Research and Safety Institute, National University of Singapore},
            country={Singapore}}

\begin{abstract}
Advances in Deep Learning bring further investigation into credibility and robustness, especially for safety-critical engineering applications such as the nuclear industry. The key challenges include the availability of data set (often scarce and sparse) and insufficient consideration of the uncertainty in the data, model, and prediction. This paper therefore presents a 
meta-learning based approach that is both uncertainty- and prior knowledge-informed, aiming at trustful predictions of material properties for the nuclear reactor design. It is suited for robust learning under limited data. Uncertainty has been accounted for where a distribution of predictor functions are produced for extrapolation. Results suggest it achieves superior performance than existing empirical methods in rupture life prediction, a case which is typically under a small data regime. While demonstrated herein with rupture properties, this learning approach is transferable to solve similar problems of data scarcity across the nuclear industry. It is of great importance to boosting the AI analytics in the nuclear industry by proving the applicability and robustness while providing tools that can be trusted. 
\end{abstract}




\end{frontmatter}


\section{Introduction}


Deep Learning introduces abundant innovative perspectives and data analytics in the physical engineering domain \citep{chen2023missing}. In the mean time, these advances also bring further investigation into credibility and robustness, especially for safety-critical engineering applications such as the nuclear industry. The key challenges include the availability of sparse and noisy data set as well as insufficient consideration of the uncertainty in the data, model, and prediction \citep{tolo2019robust}. 
It has been found that AI analytics are not widely implemented within the nuclear sector and that the sector is currently lagging behind in the Industry 4.0 revolution compared to other industries such as automotive and manufacturing \citep{prinja2022artificial}. Trustful AI techniques can be expected to play a potential role in devising robust ways of design, construction, operation, and decommission. 

To ensure the safety and effective risk management of nuclear designs, characterisation of the material properties play a significant role. Notably, the creep behavior of materials is of great concern when designing and evaluating materials for use in high-stress or high-temperature environments. 
Conventionally, time-temperature parameter (TTP) are statistically formulated using short-term data to predict the long-term rupture life, such as Orr-Sherby-Dorn (OSD) and LarsonMiller (LM) parameters, see \cite{sattar2022limitations} for a review. In recent years, the costly nature of experimental campaigns further lead to a number of machine learning algorithms leveraged as surrogate models to predict the creep rupture life \citep{biswas2020prediction, chai2023machine, wang2022high, zhang2021deep}. These models, especially neural networks, work on the principle of the universal function approximation theorem to map the nonlinear relationship between key factors in a high-dimensional space.

Essentially these models aim to predict a long-term creep rupture life based on extrapolation from the empirical short-term creep rupture data obtained experimentally. But insufficient attention are paid to address the uncertainty as often times the predictions are poorly constrained by the scarcity of data and extrapolating beyond the data. The distribution of creep data in terms of various data ranges is often highly sparse and unbalanced \citep{zhou2024creep, sattar2022limitations}. Transfer learning techniques \citep{zhou2024creep} and data augmentation procedures \citep{lye2022probabilistic} have been previously proposed and seen some mitigating effects on the limited data conundrum. But the challenge of uncertainty estimation remains.
Particularly, in this analysis, we aim to adopt a meta-learning approach that provides a generalised solution to small-sample problems while account for the uncertainty, based on explicitly learning data-driven priors from previous experience.
edo
\section{A meta-learning perspective to cross-material insights}

From a probabilistic perspective, in training a predictive model, supervised learning models aim to conduct the optimisation based on the empirical loss (i.e. likelihood):

\begin{equation}
    \theta^{*} = \arg \max_{\theta} [\mathbb{E}_{\mathcal{B}} (\sum_{x,y \in \mathcal{B}} \log p_{\theta }(y|x))]
\end{equation}

where $\mathcal{B}$ represents the batches of data pair $(x_{i}, y_{i})$ and $\theta^{*}$ denotes the optimal model parameters. However, despite abundant successes in data-rich domains such as computer vision \citep{lecun2015deep}, there remain challenges for supervised systems in the face of data scarcity, which is indeed a common issue across many physical, environmental and engineering domains \citep{chenphysics}. Generally, limited amount of data has restricted machine learning models from effectively learning
the true underlying data generating process. Significant uncertainties may exist on the model configurations that have explained the limited data and therefore the downstream extrapolation \citep{ghahramani2015probabilistic, chen2022uncertainty}. 

Prior knowledge could be  an important source of information to be employed to mitigate the impacts of limited data and efficiently guide the training and inference \citep{beer2013imprecise}. Notably, rather than subjectively eliciting priors, meta-learning involves explicitly learning priors from previous experience that lead to efficient downstream adaptation with small samples. The goal is to utilise the knowledge learned from similar tasks and adapt to another one fastly with only a few observations available. 
A key focus is the capability for fast adaptation on a new task with only limited number of data points. 

A probabilistic interpretation of meta learning indicates its two step procedures where prior information from a set of related tasks are firstly encapsulated in the meta parameters $\theta^{*} = \arg \max p(\theta^{*}|\mathcal{D}_{mt})$ (see Eq.~(\ref{eq:meta_train})), whereby a new task, often with a small number of data samples $\mathcal{C}$, can be efficiently learnt to infer the predictive distribution $p(y|x, \mathcal{C})$ for all the target data points at one time. These two stages are respectively referred to as meta-training and meta-testing.

\begin{equation}
    \theta^{*} = \arg \max_{\theta} \big[ \mathbb{E}_{\mathcal{D}_{mt}} [\mathbb{E}_{\mathcal{B}} (\sum_{x,y \in \mathcal{B}} \log p_{\theta}(y|x, C))] \big ]
    \label{eq:meta_train}
\end{equation}

where $\mathcal{D}_{mt}$ represents the meta-training set, which comprises of a set of datasets (i.e. tasks). $C$ denotes the context sets.
While meta learning generally allows to incorporate data from related tasks, another desired property is to characterise the uncertainty during training and inference. Specifically, conditional neural process (CNP) represents a new meta-learning methodology that is able to produce a distribution of predictor functions, $p(\boldsymbol{y}_{\tau}|\boldsymbol{x}_{\tau}, C)$ given the context sets, where $\tau$ denotes the target locations of interests to query at. From an architecture point of view, an encoder-decoder structure of deep neural network is built to learn a mapping from a set of datasets to predictive stochastic processes, in a meta-learning manner. 

\begin{equation}
    \boldsymbol{r} = e [h_{\theta}(\mathcal{C})]
\end{equation}

A global representation $\boldsymbol{r}$ of all the data in the context set, i.e. $ (x_{i}, y_{i}) \in \mathcal{C}$,  are encoded via the Encoder model $h_{\theta}(x_{i}, y_{i})$. Particularly, a permutation invariant operation $e$ is adopted to produce an overall summarisation.

\begin{equation}
    \phi = g_{\theta}(\boldsymbol{x}_{\tau},  \boldsymbol{r})
\end{equation}

where $\phi_{t} = (\mu_{t}, \sigma^2_{t})$ denotes a Gaussian distribution for every query data point in the target set $\boldsymbol{x}_{\tau}$. The decoder model therefore $g_{\theta}$ factorises the predictive distribution: 

\begin{equation}
    p(\boldsymbol{y}_{\tau}|\boldsymbol{x}_{\tau}, C) = \prod_{t=1}^{\tau} \mathcal{N} (y_{t}; \mu_{t}, \sigma^2_{t})
\end{equation}

It thus combines both the expressive power of multi-layer perceptron networks during training and the uncertainty characterisation capacity of Gaussian Processes during inference. 


\begin{figure*}[h!]
     \centering
     \begin{subfigure}[b]{0.49\textwidth}
         \centering
         \includegraphics[width=\textwidth]{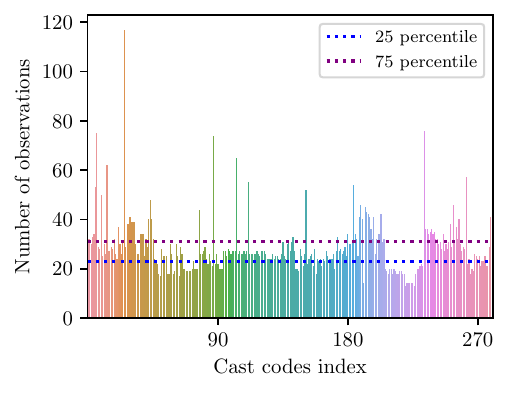}
         \caption{The distribution cast codes (tasks)}
         \label{fig:dist_cc}
     \end{subfigure}
    \hspace{0pt}
     \begin{subfigure}[b]{0.49\textwidth}
         \centering
         \includegraphics[width=\textwidth]{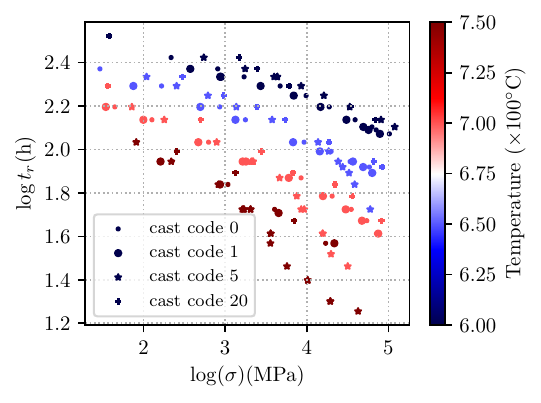}
         \caption{Some example cast codes}
         \label{fig:ex_ccs}
     \end{subfigure}
\end{figure*}

\section{Numerical experiments on creep properties}

A database of the material properties for 58 different steel types is obtained from the National Institute for Materials Science (NIMS) based on a previous experimental campaign under the Material Properties Predictor for Power Plant Steels (M4PS) project \citep{prinja2022artificial}. This includes 8005 experimental observations on the creep rupture properties in terms of stress, temperature, fracture time, etc.
Fig.~\ref{fig:dist_cc} shows a distribution the number of 
observations with respect to various cast codes. There are 44 materials associated with 281 cast codes and also 35 temperature levels involved in the experiments. It shows that most cast codes are associated with less than 40 observations, and they will be much sparser if factored in the temperature distributions. Particularly, there exists quite a few cast codes that only has less than 20 experimental observations. As a concrete example, Fig.~\ref{fig:ex_ccs} displays the data for some randomly selected cast codes as an illustration. Based on these samples, it is not hard to spot a plausible linear trend between the stress and creep rupture time, both in logarithmic scale, when group by temperature.

In this analysis, we take on a meta-learning approach to quickly and robustly learn an uncertainty-aware model to predict the creep rupture life of materials with different cast codes given only a very limited number of experiments. It should be noted that our aim is for a generalised meta-model that applies to any arbitrary cast code with limited observations, as opposed to only targeting specifically at a certain material or cast code. It serves as a widely useful approach for robust learning of machine learning models given limited data and express its uncertainty.

To demonstrate the performance of learning given data scarcity, the model has been specifically tested on the a held-out set of 20 cast codes with the least number of observations, whose quantity of observations range from 13 to 18. The testing configuration entails that, for a certain cast code,  only several samples (entitled as context points) are available to the model. The learned model will be further tested on separate data points (referred to as target points). It would be extremely hard for traditional supervised learning paradigms to learn effectively without overfitting given such small number of samples.

The prior knowledge has been learnt during the meta-training stage where the data of a number of 261 cast codes are involved in the training and validation. A proportion of 20\% of the training data are randomly taken as the validation set where the hyperparameters of model are optimised.
Meanwhile, the model is compared with the baseline of Larson-Miller model with degrees of $d=1$ and $d=2$. 
The Larson-Miller relations is one of the most common classic techniques in representing creep-rupture data. By characterising a time-temperature parameter (i.e. $P_{LM}$ the Larson-Miller parameter), the Larson–Miller model (LM), shown at Eq.~(\ref{eq:LM}) below, predicts the lifetime of material with respect to time and temperature using a correlative approach based on the Arrhenius rate equation \citep{maruyama2018physical}.

\begin{equation}
    P_{LM} = \log(t_{r} + C_{LM})T_{r} / 1000 = h_{\xi}(\log \sigma)
    \label{eq:LM}
\end{equation}

where $T_{r}$ denotes temperature in the unit of Rakine; $t_{r}$ denotes the stress-rupture time; and $C_{LM}$ is a constant which is often assumed to takes the value of 20. $h_{\xi}(\log \sigma)$ denotes the hypothesis between the Larson-Miller parameter and the stress in log scale, typically in the form of a linear function of polynomial basis parameterised by $ \xi$ \citep{zhang2023method}.

\begin{equation}
    h_{\xi}(\log \sigma) = {\xi}^{T} \phi(\log \sigma)
    \label{eq:LM}
\end{equation}

where $\phi_{j}(x)=x^{j}$ suggests a polynomial basis. It should be noted that Eq.~(\ref{eq:LM}) indicates a linear model with respect to parameters, but it characterises a non-linear relationship with respect to the independent variables. Particularly, it is evident that $d=1$ corresponds to a linear relationship between stress and $P_{LM}$. In this regime, the TTP (i.e. $P_{LM}$) serves as a proxy where one can derive the rupture time at a further step given temperature values. However, the trained conditional neural process model will directly yield the predictive distributions of the rupture time.

\begin{figure*}[h!]
\centering
     \begin{subfigure}[b]{0.49\textwidth}
         \centering
         \includegraphics[width=\textwidth]{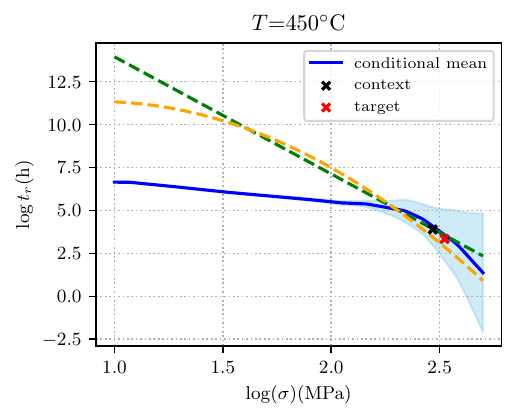}
         \caption{$T$=450 \textcelsius}
         \label{fig:T450}
     \end{subfigure}
\hspace{0pt}
     \begin{subfigure}[b]{0.49\textwidth}
         \centering
         \includegraphics[width=\textwidth]{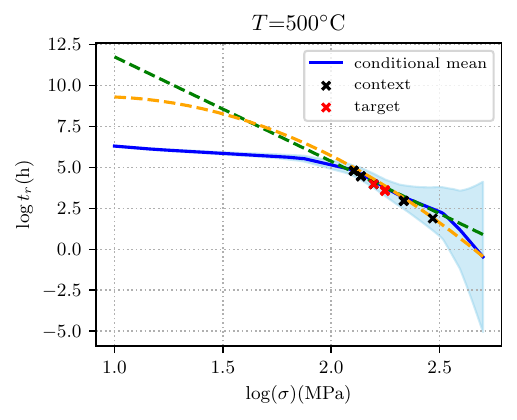}
         \caption{$T$=500 \textcelsius}
         \label{fig:T500}
     \end{subfigure}
\hspace{0pt}
     \begin{subfigure}[b]{0.49\textwidth}
         \centering
         \includegraphics[width=\textwidth]{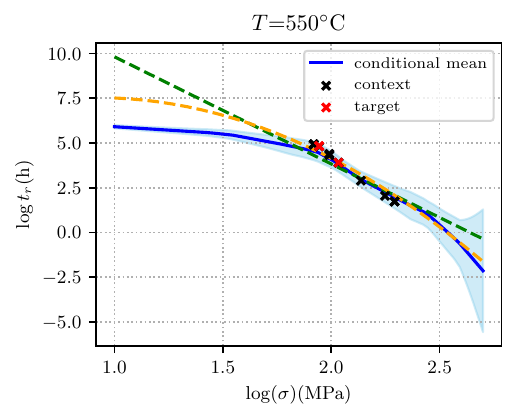}
         \caption{$T$=550 \textcelsius}
         \label{fig:T550}
     \end{subfigure}
\hspace{0pt}
     \begin{subfigure}[b]{0.49\textwidth}
         \centering
         \includegraphics[width=\textwidth]{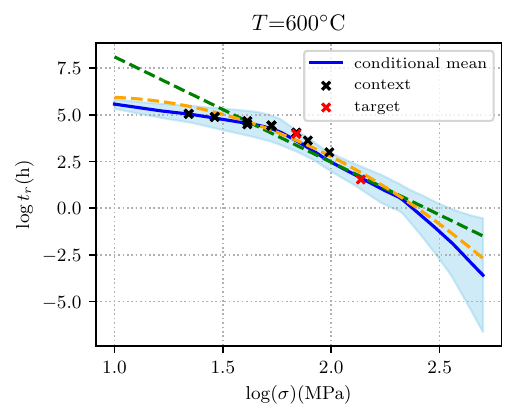}
         \caption{$T$=600 \textcelsius}
         \label{fig:T600}
     \end{subfigure}
\caption{Comparison of the trained model along with two baselines on an arbitrary cast code in the validation set. The green line suggests the LM model with $d=1$ while the yellow curve suggests the LM model with $d=2$. The blue curve denotes the conditional mean produced by the trained model and the blue shades denote the range of 2 standard deviation}
\label{fig:example_gp_like}
\end{figure*}

Fig.~\ref{fig:example_gp_like} shows the generalisation performance of the trained model under varying temperature conditions. Black crosses denote the context data samples while the red crosses denote the unseen samples for testing purposes. The sparsity of data distributions with respect to temperature can be easily seen in these figures. Often times there are very limited number of experiments available or even none for some temperature setting (e.g. merely 2 observations for $T$=450 \textcelsius $ $ and no observations for temperature higher than $T$=650 \textcelsius $ $ in this case).

Compared to the deterministic LM models, the trained conditional neural process model produces the predictive uncertainty on its extrapolation. 
It is vital to characterise the epistemic uncertainty when extrapolating to unseen data ranges. The underlying predictive model may not even be physically correct in generalising to environments outside the scope of observations. For the high stress range ($\log (\sigma) > 2.5$), both LM models in all temperature scenarios are well included within the variance of the trained model, showing a certain degree of agreement in extrapolating towards high stress conditions. The nonlinear $d=2$ case shows relatively better agreements with the conditional mean of the proposed model.
However, these models differ considerably in the the low-stress range, which is of typical interests to engineers who are finding the long-term creep behavior of materials. The proposed model manifests the lowest extrapolated estimates among the three, while the $d=1$ model gives the highest estimate due to its linear assumption. 
We note that, given unseen data ranges the proposed model yields the extrapolation based on the learned prior insights from many related tasks due to the mechanism of meta-training and may well be the most reliable extrapolation. 

\begin{figure}[h!]
     \centering
     \begin{subfigure}[b]{0.32\textwidth}
         \centering
         \includegraphics[width=\textwidth]{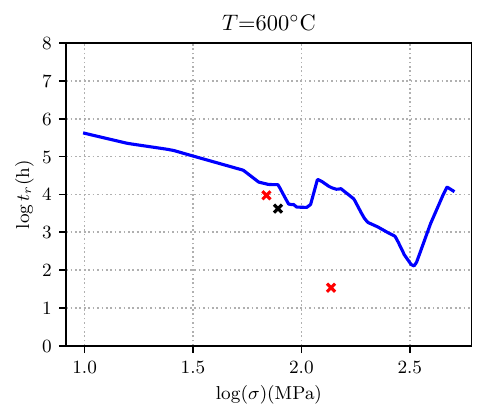}
         \label{fig:A}
     \end{subfigure}
    \hspace{0pt}
     \begin{subfigure}[b]{0.32\textwidth}
         \centering
         \includegraphics[width=\textwidth]{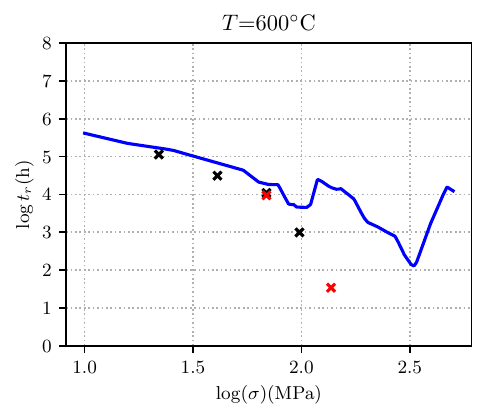}
         \label{fig:B}
     \end{subfigure}
    \hspace{0pt}
     \begin{subfigure}[b]{0.32\textwidth}
         \centering
         \includegraphics[width=\textwidth]{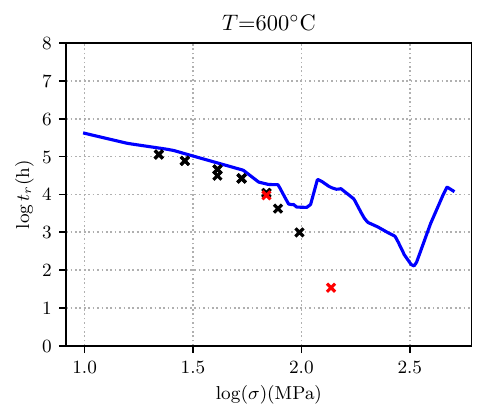}
         \label{fig:c}
     \end{subfigure}
    \medskip
    \centering
     \begin{subfigure}[b]{0.32\textwidth}
         \centering
         \includegraphics[width=\textwidth]{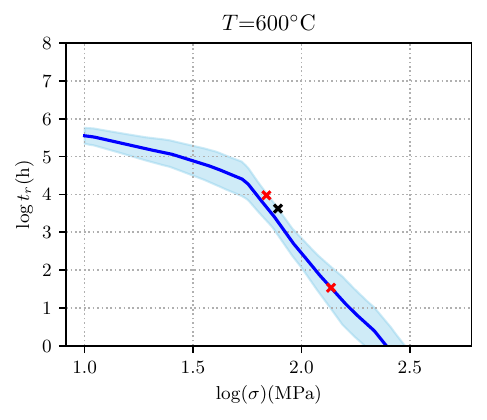}
         \label{fig:D}
     \end{subfigure}
    \hspace{0pt}
     \begin{subfigure}[b]{0.32\textwidth}
         \centering
         \includegraphics[width=\textwidth]{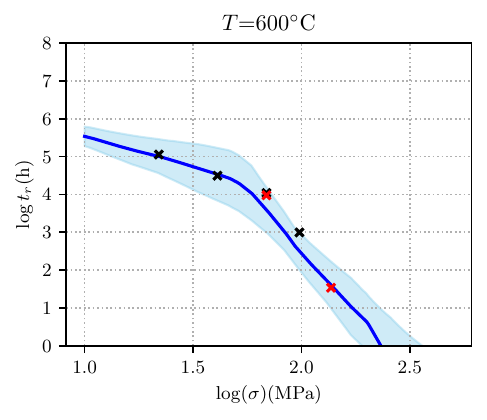}
         \label{fig:E}
     \end{subfigure}
    \hspace{0pt}
     \begin{subfigure}[b]{0.32\textwidth}
         \centering
         \includegraphics[width=\textwidth]{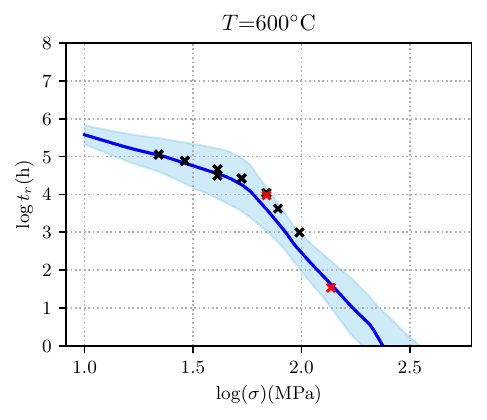}
         \label{fig:F}
     \end{subfigure}
    \medskip
    \centering
     \begin{subfigure}[b]{0.32\textwidth}
         \centering
         \includegraphics[width=\textwidth]{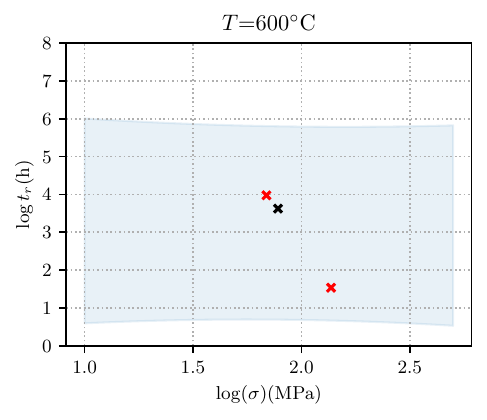}
         \caption{Few context points}
         \label{fig:G}
     \end{subfigure}
    \hspace{0pt}
     \begin{subfigure}[b]{0.32\textwidth}
         \centering
         \includegraphics[width=\textwidth]{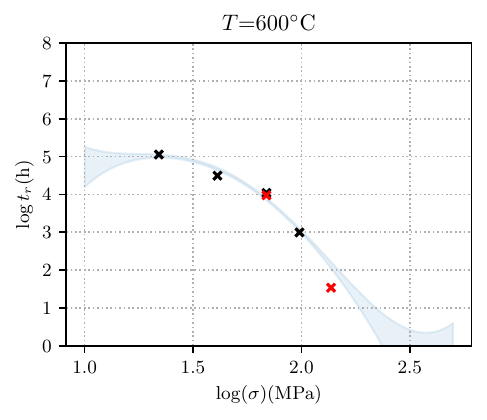}
         \caption{Some context points}
         \label{fig:H}
     \end{subfigure}
    \hspace{0pt}
     \begin{subfigure}[b]{0.32\textwidth}
         \centering
         \includegraphics[width=\textwidth]{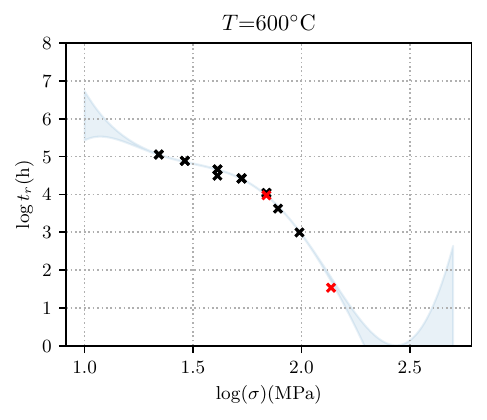}
         \caption{Rich context points}
         \label{fig:I}
     \end{subfigure}
        \caption{Performance comparison of varying context points. \textbf{top row} represents the predictions from the a pretrained neural network model; \textbf{middle row} predictions from the conditional neural process model; \textbf{bottom row} predictions from a Gaussian Process model. From the first to the last column, more context points are available to the models}
        \label{fig:varying_context_points}
\end{figure}

A demonstration can be seen from Fig~\ref{fig:varying_context_points}, where the proposed model is compared with a pretrained neural network model and a Gaussian Process model, under varying number of context points. The context set and target set are from an arbitrary cast code in the validation data set. It should be noted this validation task have observations of multiple temperature conditions and the temperature is fixed herein for plotting convenience.
The conditional neural process model is designed to (I) learn prior knowledge from the meta-learning stage; (II) yield a distribution of predictor functions to express confidence during inference; Therefore, it can be seen as a combination of both worlds, integrating with both pre-training and also the ability to leverage context points similar to a gaussian process (GP). It can be seen that an increasing number of context points have no impact on a pretrained model which merely learns from the set of meta-tasks. In this setting, these context points can practically been seen as targeting points. Despite not leveraging the context points which contains the most direct information pertaining to the task under testing, this model can still produce a reasonable extrapolation in the low stress range due to the pretraining on those related tasks. But it completely ignores the context information most relevant to the task under investigation and produces incorrect results on stress level larger than 2.
In comparison, a Gaussian process have no such pretraining and instead produce the predictive distribution completely based on the context points.  It can be seen, from the third row of Fig.~\ref{fig:varying_context_points}, that with very few context points, we can only yield the non-informative prior gaussian process. 
With some additional number of data points, the GP adapt to the context points and produce uncertainty bounds accordingly. Importantly, we can see the shift of low-stress extrapolation of the mean prediction $\log t_{r}$ from roughly 5 to 6 with just a few more observations. This shows that GP is relatively more sensitive to the context points and will probably produce different results given the observations. However, conditional neural process has both the knowledge of related tasks and also the capacity to adapt on the basis of context points. 
As shown in the second row, it produces reasonable low-stress extrapolation even with very few context points, serving as a informative starting point for further adaptation. Given more context points, it fine-tunes its predictions. 

\begin{figure}[h]
\includegraphics[width=9cm]{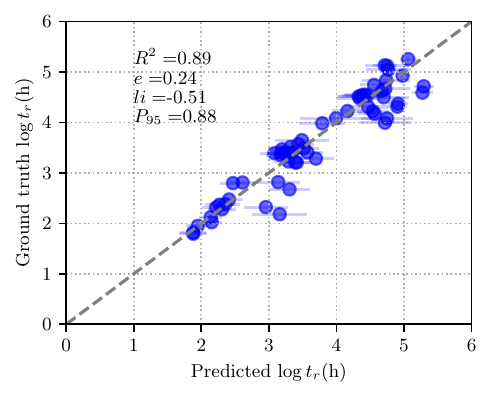}
\centering
\caption{The prediction performance of the conditional neural process model over the whole testing set. The horizontal bar represents 2 times of standard deviations}
\label{fig:all_test_tasks}
\end{figure}

Collectively, Fig.~\ref{fig:all_test_tasks} displays the prediction against the whole held-out test set which includes 60 observations across 20 cast codes with the least number of experiments. The number of experiments range from 13 to 18. Meanwhile, several metrics are proposed to showcase the generalisation performance. $e$ denotes the mean absolute error and $li$ represents the log likelihood in the testing set, these two reflect the quality of the model fit. On the other hand, $P_{95}$,  as an uncertainty measure, evaluates the ratio of the ground truth to be captured by the predicted interval.

\begin{equation}
    e = \frac{1}{n} \sum_{n=1}^{n} | y_{i} - \hat{y_{i}} |
    \label{eq:mae}
\end{equation}

\begin{equation}
    li = \log p(\boldsymbol{y}_{\tau}|\boldsymbol{x}_{\tau}) 
\end{equation}

\begin{equation}
    P_{95} = c_{t} / n_{t},\  \text{with} \ c_{t} = \sum_{i=1}^{n} c_{i}
\end{equation}

\begin{equation}
    c_{i} = \begin{cases}
 0, y_{i} \in [y_{U_{i}}, y_{L_{i}}] \\ 
 1, y_{i} \notin [y_{U_{i}}, y_{L_{i}}]
\end{cases}
\end{equation}

where $c_{t}$ is defined by a vector of length $n_{t}$ (total number predictions), whose element $c_{i}$ indexes a true value captured by the estimated credible interval. 
Considering the situation of prediction under very limited data, Fig.~\ref{fig:all_test_tasks} shows great performance with the $R^{2}=0.89$ and $P_{95}=0.88$, indicating 88\% of the predicted intervals including the ground truth.
To comprehensively showcase the generalisation performance and consider randomness, Table~\ref{table:testing_performance} tabulates the metrics across 20 random runs on the whole testing set. Both $li$ and $P_{95}$ provide a basis for comparing with other probabilistic methods. It can be seen that, besides being more informative than the LM baselines (with additional uncertainty measures), the proposed model is also more accurate. Meanwhile, for the LM models, the accuracy is decreasing as with higher degrees of the polynomial, which suggests the sign of overfitting when relating the stress and the TTP. Such overfitting is expected due to the small number of observations.

\begin{table}[h!]
\centering
\caption{Performance metrics on the prediction of rupture time
for the testing tasks with the least number of observations over 20 random runs}
\label{table:testing_performance}
\begin{tabular}{@{}ccccc@{}}
\toprule
\multirow{2}{*}{} & \multirow{2}{*}{proposed model} & \multicolumn{3}{c}{Larson Miller relation} \\ \cmidrule(l){3-5} 
                  &                                         & $d$=1          & $d$=2          & $d$=3          \\ \midrule
$e$                 & 0.261 $\pm$ 0.026                     & 0.354 $\pm$ 0.105 & 0.375 $\pm$ 0.183 & 0.449 $\pm$ 0.333 \\
$li$                & -0.409 $\pm$ 0.125                    & -             & -             & -             \\ 
$P_{95}$            & 0.878 $\pm$ 0.040                     & -             & -             & -             \\ 
\bottomrule
\end{tabular}
\end{table}

\section{Conclusion}

In this paper a meta-learning based approach is developed for robust creep rupture life prediction for nuclear reactor design, which typically suffers from significant epistemic uncertainty due to extrapolation and having very limited observation when building models. Insufficient attention are paid to uncertainty characterisation despite some attempts of adopting machine learning techniques.  
The proposed approach demonstrates superior generalisation performance than conventional baseline methods with higher accuracy and provides the necessary confidence associated with such prediction. This superior performance is mostly driven by its two vital capabilities: (I) extracting prior knowledge from related tasks and quickly adapt to new tasks; (II) yielding uncertainty-aware extrapolation through a distribution of predictor functions.

The proposed technology allows considering the unavoidable variability associated with any experimental campaign required to characterise materials. The impact of this analysis will prove the transferable applicability of probabilistic AI analytics in the nuclear industry by providing tools that can be trusted.

\section{Acknowledgement}

This work was supported Enhanced Methodologies for Advanced Nuclear System Safety(eMEANSS) [Project no. EP/T016329/1]. Data can be found publically at \url{https://github.com/Adolphus8/Project_PROMAP}.

\bibliographystyle{elsarticle-harv} 
\bibliography{cas-refs}





\end{document}